\pdfoutput=1

\documentclass[11pt]{article}

\usepackage[final]{acl}

\usepackage{times}
\usepackage{latexsym}

\usepackage[T1]{fontenc}

\usepackage[utf8]{inputenc}

\usepackage{microtype}

\usepackage{inconsolata}

\usepackage{url}
\usepackage{xcolor}
\usepackage{graphicx}
\usepackage{tabularx}
\usepackage{array}
\usepackage{amsmath}
\usepackage{amssymb}
\usepackage{setspace}
\usepackage{multirow}
\usepackage{makecell}
\usepackage{colortbl}
\usepackage{hhline}
\usepackage{booktabs}
\usepackage{dcolumn}
\usepackage{caption}
\usepackage{subcaption}
\usepackage{chngpage}
\usepackage{ctable}
\usepackage{diagbox}

\usepackage{enumitem}
\title{Modeling Layout Reading Order as Ordering Relations for Visually-rich Document Understanding}

\author{
    Chong Zhang$^{1,2}$\thanks{\ \ Part of work is done during an internship at Ant Group.}, 
    Yi Tu$^{3}$, Yixi Zhao$^{1,2}$, Chenshu Yuan$^{4}$, Huan Chen$^{3}$, Yue Zhang$^{1,2}$, \\
    \textbf{Mingxu Chai}$^{1,2}$, \textbf{Ya Guo}$^{3}$, \textbf{Huijia Zhu}$^3$, \textbf{Qi Zhang}$^{1,2}$\thanks{\ \ Corresponding author.}, \textbf{Tao Gui}$^{2,5\dagger}$ \\
    $^1$School of Computer Science, Fudan University, Shanghai, China \\ 
    $^2$Shanghai Key Laboratory of Intelligent Information Processing, Shanghai, China \\
    $^3$Ant Tiansuan Security Lab, Ant Group, Hangzhou, China \\
    $^4$School of Statistics and Data Science, Nankai University, Tianjin, China \\
    $^5$Institute of Modern Languages and Linguistics, Fudan University, Shanghai, China \\
    \texttt{\{chongzhang20, qz, tgui\}@fudan.edu.cn, \{qianyi.ty,guoya.gy\}@antgroup.com} \\
}

\begin{document}
\maketitle
\begin{abstract}
Modeling and leveraging layout reading order in visually-rich documents (VrDs) is critical in document intelligence as it captures the rich structure semantics within documents. 
Previous works typically formulated layout reading order as a permutation of layout elements, i.e. a sequence containing all the layout elements. 
However, we argue that this formulation does not adequately convey the complete reading order information in the layout, which may potentially lead to performance decline in downstream VrD tasks. 
To address this issue, we propose to model the layout reading order as ordering relations over the set of layout elements, which have sufficient expressive capability for the complete reading order information. 
To enable empirical evaluation on methods towards the improved form of reading order prediction (ROP), we establish a comprehensive benchmark dataset including the reading order annotation as relations over layout elements, together with a relation-extraction-based method that outperforms previous methods. 
Moreover, to highlight the practical benefits of introducing the improved form of layout reading order, we propose a reading-order-relation-enhancing pipeline to improve model performance on any arbitrary VrD task by introducing additional reading order relation inputs. 
Comprehensive results demonstrate that the pipeline generally benefits downstream VrD tasks: 
(1) with utilizing the reading order relation information, the enhanced downstream models achieve SOTA results on both two task settings of the targeted dataset; 
(2) with utilizing the pseudo reading order information generated by the proposed ROP model, the performance of the enhanced models has improved across all three models and eight cross-domain VrD-IE/QA task settings without targeted optimization. 
\end{abstract}


\section{Introduction}

\begin{figure}[t]
    \centering
    \includegraphics[width=\linewidth]{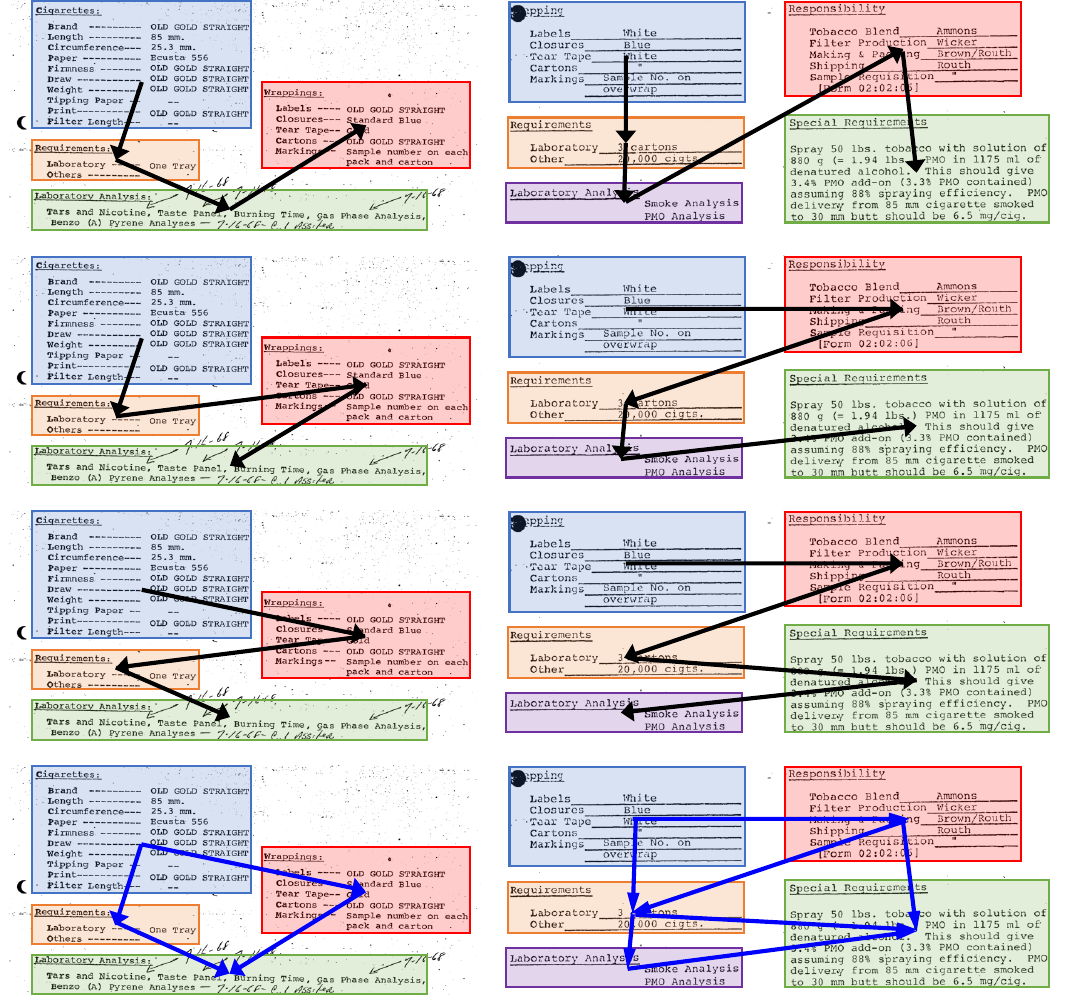}        
    \caption{
        Motivation of reformulating layout reading order. 
        In complex document layouts, multiple reading sequences are acceptable (displayed in the first three rows); thus the reading order information may be incomplete if represented by one single sequence. 
        We propose to represent the relationship of \textit{immediate succession during reading} among layout elements using a directed acyclic relation (displayed in the last row as a directed acyclic graph), ensuring that the complete layout reading order information is conveyed. 
        }
    \vspace{-4.5mm}
    \label{fig:page1}
\end{figure}

\begin{figure*}[t]
    \centering
    \vspace{-5mm}
    \includegraphics[width=0.99\linewidth]{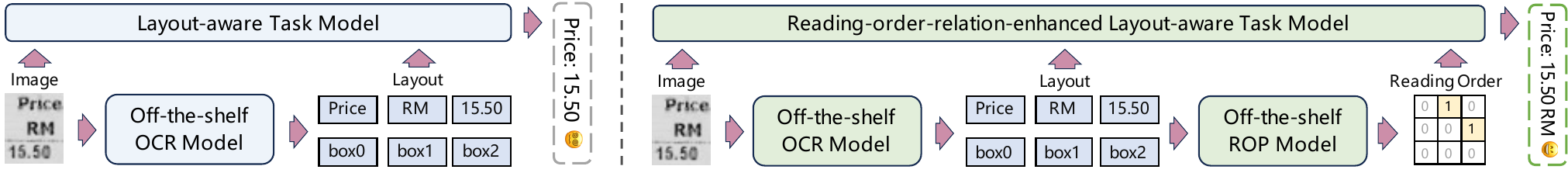}   
    \vspace{-1mm}
    \caption{
        The reading-order-relation-enhancing pipeline (right, green) comparing with the original pipeline (left, blue) for general document processing. 
        "RM" denotes Malaysian Ringgit. 
        }
    \vspace{-4mm}
    \label{fig:rore}
\end{figure*}

The field of visually-rich documents (VrDs) has attracted widespread attention recently. 
Towards the growing industrial demands of automated processing of VrDs, a wide range of challenging tasks have arisen including understanding, classifying, extracting, and post-processing the information of VrDs \citep{rvl-cdip-harley2015evaluation,huang2019icdar2019,wang2021layoutreader,10.1145/3580305.3599929,10.1007/978-981-99-7894-6_10}.
Among those tasks, Reading Order Prediction (ROP) is a peculiar task on VrD layouts aiming to determine the human-acceptable reading order of layout elements, i.e. 2D-situated words or text regions (segments) given their bounding boxes. 
ROP is a fundamental task for layout understanding as it reveals the influence of layouts for human reading strategies \citep{wright1999psychology,dyson2004physical}. 
The task is challenging due to the diversity and complexity of layout designs among documents, and is also important in applications as previous works have demonstrated that the performance of downstream VrD tasks can be substantially improved by leveraging proper reading order information \citep{peng2022ernie,zhang2023reading}.

However, we argue that the prevailing task formulation of ROP is inappropriate as it does not convey the complete layout reading order information. 
In recent works, the layout reading order is typically represented as a permutation of layout elements, i.e. a sequence containing all the words or segments within the layout \citep{malerba2007learning,li2020end}.
However, as shown in Fig. \ref{fig:page1}, there exists multiple permutations of the displayed layout that represents a human-acceptable reading order, and some of which conflict with other ones. \footnote{In Fig. \ref{fig:page1}, annotations are drawn as block-level for better visualization, which is different from the proposed dataset. }
Therefore, \textbf{representing the layout reading order information with one single permutation is unable to convey the complete reading order information in the layout}, and could possibly inject noise in the annotations.
Both issues negatively affects its application in downstream tasks. 
Given the problem, we propose two questions:
(1) How to improve the formulation of reading order to include the complete information?
(2) Would the improved formulation be truly helpful in practical applications that downstream VrD tasks could be further benefited by leveraging the improved form of reading order information?

To address the first question, we propose to model the layout reading order as ordering relations over the set of layout elements. 
Since there lacks a clear definition of layout reading order, we investigate its properties to conceptualize it with two terms, each of which is represented as an ordering relation: (1) Immediate Succession During Reading (ISDR) is modeled as a directed acyclic relation over layout elements; (2) Generalized Succession During Reading (GSDR) is modeled as the transitive closure of ISDR, which is a strict partial order relation over layout elements. 
ROP is then reformulated as a relation extraction task on layout elements.
For developing automatic methods of ROP in the improved form, we also establish a new benchmark dataset annotating ISDR as relation pairs of layout segments together with a baseline relation-extraction-based method. 
Experiments demonstrate that the proposed method outperforms previous sequence-based methods by a large margin on ROP. 

To address the second question, we propose the reading-order-relation-enhancing (RORE) pipeline which is confirmed to be effective by comprehensive experiments on downstream VrD-IE/QA tasks. 
For injecting reading order relation information, we propose to represent it in a $n*n$ binary matrix over $n$ input textual tokens of PTLMs, and introduce a relation-aware attention mechanism to incorporate the reading order information. 
Experiment results show that with additional reading order inputs, the performance of three baseline PTLMs have shown universal improvement.
Moreover, considering that ground truth annotation of reading order information is unavailable in real-world scenarios, it is expected to leverage pseudo reading order labels generated by an off-the-shelf ROP model in any arbitrary downstream tasks. 
The expectation is proven feasible by experiment results, since the performance of three reading-order-relation-enhanced models are comprehensively improved across eight prevailing benchmarks of VrD-IE/QA from different domains, achieving SOTA on two of them without targeted optimization.

\begin{figure*}[t]
    \centering
    \vspace{-5mm}
    \includegraphics[width=0.9\linewidth]{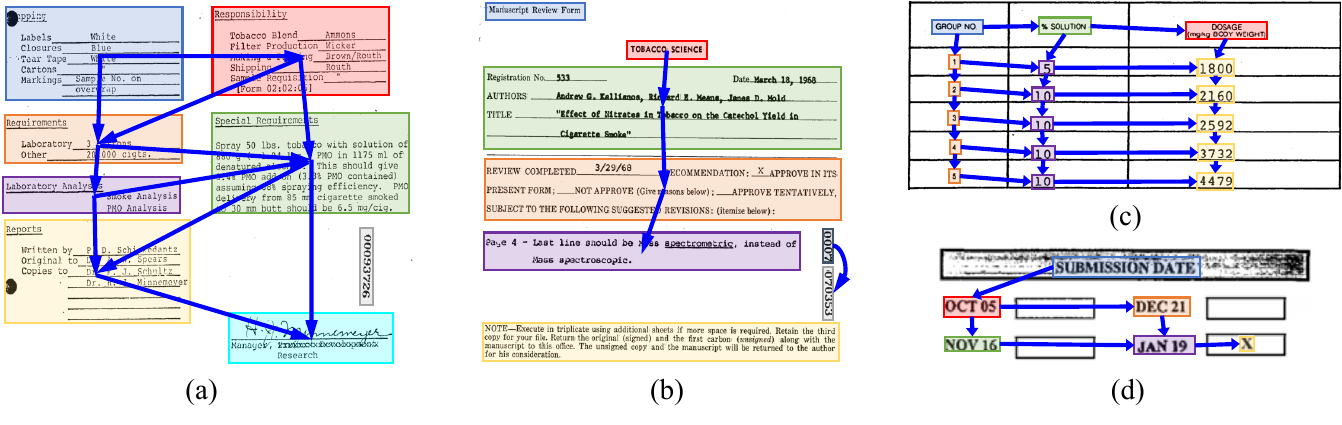}   
    \vspace{-4mm}
    \caption{
        Several example layouts with non-linear reading order.
        Annotations are drawn as block-level for better visualization. 
        (a) The complex layout includes multiple possible reading sequences (illustrated in Fig. \ref{fig:page1});
        (b) The reading order of header, footer and watermark within the layout are separated from the main body;
        (c) The table within the layout can be read either vertically or horizontally;
        (d) Indirect reading order relationship is also important as relevant elements may be separated by other contents. 
        }
    \vspace{-4mm}
    \label{fig:motivation}
\end{figure*}

\noindent Overall, the main contributions of this work are:
\begin{enumerate}[leftmargin=*,noitemsep,topsep=0pt]
    \item 
    We identify the existing problem that the current task form of ROP fails to convey the complete layout reading order information, and suggest modeling the reading order as ordering relations over layout elements. 
    We introduce a new comprehensive benchmark for quantitative evaluation of the improved formulation of ROP. 
    \item 
    We propose a relation-extraction-based model which outperforms previous sequence-based methods on ROP.
    Further, the proposed model generate high-quality pseudo reading-order-relation labels for arbitrary inputs from various domains. 
    \item 
    We introduce a pipeline to enhance arbitrary VrD tasks based on the proposed ROP model. 
    Comprehensive results demonstrate that the performance of several prevailing methods on various downstream VrD-IE/QA tasks can be universally improved by leveraging the (probably pseudo) reading order relation information. The enhanced methods have achieved SOTA results on the targeted benchmark and also two cross-domain task settings.
\end{enumerate}
The proposed dataset and code implementation of proposed models are released at \url{https://github.com/chongzhangFDU/ROOR}.

\section{Reading Order as Ordering Relations}

\subsection{Conceptualization of Reading Order}
\label{sec:isdr}

This section conceptualizes the reading order of layouts by examining general human reading patterns. 
While layout reading order prediction is one of the most crucial task in document AI, the concept of reading order itself is rarely discussed and has not been fully clarified. 
We define the term \textbf{Immediate Succession During Reading} (ISDR) to precisely capture the human reading order on general textual contents.

Intuitively, the layout reading order is a manifestation of human reading strategies tackling different layout designs \citep{wright1999psychology,dyson2004physical}.
To precisely define the layout reading order, a formalized description of human reading behavior on layouts is required.
For this purpose, we draw inspiration from human behavior of reading plain text.
Plain text is always a verbatim sequence following its inherent order, where each word appears after the preceding one linearly. 
We refer to each word as the immediate successor of its preceding word, since it is read immediately after the preceding word during reading.
In layouts, when we read a layout element subsequently after the previous element, we call the layout element as the immediate succession of the previous one. 
Thereby, we introduce the term \textit{immediate succession during reading} to describe the human reading order of general textual contents. 

\subsection{ISDR as a Directed Acyclic Relation}
\label{sec:ro_as_or}
In this section, we illustrate that current task form of reading order prediction on layout is not suitable by examining the difference between human reading habits of plain text and layout.
To address the problem, we suggest representing the ISDR relationship as a directed acyclic relation (DAR, formal definition in Appendix \ref{sec:theory}) over layout elements, and reformulate the ROP task of layout as a relation extraction task.

As illustrated in \S\ref{sec:isdr}, the reading order of plain text can be naturally represented as a permutation of words, i.e. the word sequence itself. 
Most previous works of layout reading order prediction have followed this manner and also represent the layout reading order as a permutation of layout elements. 
However, this formulation is unsuitable due to the following characteristics of layout: 
(1) Different to plain text where words are succeeded by one single subsequent word, layout elements within a document can have multiple immediate successors (Fig. \ref{fig:motivation}a and \ref{fig:motivation}c), and table cells can have successors to the right or below (Fig. \ref{fig:motivation}c). 
(2) Some certain layout elements do not interact with others, and their successive relationships should also be separated from others (Fig. \ref{fig:motivation}b). 

To address these problems, we suggest representing the ISDR relationship as a DAR over layout elements as an alternative formulation. 
Motivated by Fig. \ref{fig:motivation} which portrays ISDR as a directed graph of layout elements, we suggest conceptualizing ISDR as a binary relation over the set of layout elements.
Given that ISDR represents a succession relationship, we further stipulate that this binary relation is acyclic to prevent any circular dependencies of layout elements.
As a result, ISDR is formulated as a DAR over the layout elements.

\subsection{Generalized Succession During Reading}
\label{sec:gsdr}

In this section, we propose Generalized Succession During Reading (GSDR) as an alternative form of reading order information.
GSDR focuses on the global awareness of reading order relation between layout elements, and is formulated as a strict partial order (SPO, formal definition in Appendix \ref{sec:theory}) over the set of layout elements.

The ISDR relationship proposed in \S\ref{sec:isdr} conveys the complete reading order information, and it mainly focuses on the local reading order relationship between layout element. 
However, there are cases in reading where we focus on the global reading order information.
For example, as illustrated in Fig. \ref{fig:motivation}d, the question {\footnotesize \textit{"SUBMISSION DATE"}} is always read before the answer {\footnotesize \textit{"SEP 15"}}, yet their reading order relationship is not directly represented by ISDR as the answer does not directly follow the question during reading. 
%
This motivates us to introduce the transitive closure of ISDR, namely \textit{generalized succession during reading}, to convey the global subsequent relationship between layout segments. 
Specifically, an element being the generalized successor of another one indicates that this element is read after another element somehow in reading. 
From the order-theoretic perspective, the GSDR relationship over layout elements is a SPO, different from the GSDR relationship over words within plain text as a strict total order, indicating the different characteristics of plain text and layout. 
The application of GSDR is discussed in Appendix \ref{sec:ablation}. 

\subsection{Dataset Construction}

In this section, we present \textbf{ROOR} (\textbf{R}eading \textbf{O}rder as \textbf{O}rdering \textbf{R}elations), a benchmark dataset featuring segment-level reading order relation annotations. 
This benchmark is constructed to facilitate the development of automated ROP methods of the improved task form. 

The annotation of ROOR is based on EC-FUNSD \citep{zhang2024rethinking} due to its variety of document layouts and high-quality layout annotations. 
In EC-FUNSD, each sample is a single-page scanned VrD with manually-reviewed layout annotations emulating the standard output of word-and-segment annotations produced by OCR engines.
Two domain experts independently annotated the ISDR relation as linking pairs of segments in each document, and resolved any conflicts between their annotations through discussion. 
To maintain the property of ISDR relationship as a DAR in each sample, the line-breaks within grid-lined tables were not annotated, i.e. the succession from the last cell in the previous row to the first cell in the next row. The same was applied to column-breaks.
Overall, ROOR comprises 199 samples including 10,662 segments, 31,297 words and 10,967 annotatoed reading order linking pairs.

Statistics on ROOR indicate the prevalent occurrence of non-linear reading order within layouts. 
We define a segment as involved in non-linear reading order iff. both its preceding and succeeding segment uniquely exist (except for the initial and final segments which naturally lack either a predecessor or successor). 
Examination on ROOR shows that 23.76\% of segments within the dataset are involved in non-linear reading order, demonstrating the significance to deal with this phenomenon in layout processing.

\section{Methodologies}

\subsection{Reading Order Prediction as Relation Extraction}

As we model the layout reading order as ordering relations, the reading order prediction task is then reformulated as a relation extraction task as described below:
A document layout with $N_{\mathcal{D}}$ layout elements (i.e. words or segments) is represented as $\mathcal{D}=\{(w_i, \textbf{b}_i)\}_{i=1,\dots,N_{\mathcal{D}}}$, where $w_i$ denotes the $i$-th element in document and $\textbf{b}_i=(x_i^0, y_i^0, x_i^1, y_i^1)$ denotes the position of $w_i$ in the document layout. The coordinates $(x_i^0, y_i^0)$ and $(x_i^1, y_i^1)$ correspond to the bottom-left and top-right vertex of $w_i$'s bounding box, respectively. 
The set of reading order relation pair of elements from $s_{\mathcal{D}}$ is denoted as $p_{\mathcal{D}} = \{p_1,\dots,p_K\}$, where the $k$-th relation pair $p_k=(ks,ko)$ indicates that $w_{ko}$ is an immediate successor of $w_{ks}$, ($1 \leq ks, ko \leq N_{\mathcal{D}}$). 
The aim of ROP is to recognize all the reading order relation pairs within the document. 

\subsection{A Baseline Model for ROP}

After treating the task as relation extraction on VrD layouts, a baseline model is then introduced. 
To avoid over-engineering for the task, we straightforwardly adopt the global pointer network \citep{su2022global}, a widely used architecture for relation extraction tasks in NLP. 

For a document layout $\mathcal{D}$ with $N$ layout elements, let $L$ be the ground truth label of reading order relation, where $(i, j) \in L$ indicates that the $j$-th element is an immediate successor of the $i$-th element.
The tokenized input of this document is denoted as $(x_{1}^1,\dots,x_{1}^{n_1},\dots,x_{N}^1,\dots,x_{N}^{n_N})$, where $x_{i}^j$ indicates the $j$-th token of the $i$-th layout element. 
Let $\textbf{b}_{i}^j$ be the corresponding bounding box of $x_{i}^j$, the tokenized input is then fed to the backbone encoder to obtain the layout-aware text embeddings: 
\begin{equation}
\resizebox{\linewidth}{!}{$
h_{1}^1,\dots,h_{N}^{n_N} = {\rm{PTLM}}(x_{1}^1,\dots,x_{N}^{n_N}; \textbf{b}_{1}^1,\dots,\textbf{b}_{N}^{n_N})
$}
\end{equation}
where 
${\rm{PTLM}}$ is the encoder of text and layout inputs;
$h_{i}^j$ is the layout-aware embedding of $x_{i}^j$.

\noindent Layout element embeddings are then calculated by gathering their corresponding token embeddings:
\begin{equation}
h_{i} = {\rm{AvgPool1d}}(h_{i}^1,\dots,h_{i}^{n_i})
\end{equation}
where $h_{i}$ is the embedding of the $i$-th element.

\noindent Then the embeddings are fed into the global pointer network for relation extraction:
\begin{equation}
\begin{array}{l@{}l}
q_i = \textbf{W}_q h_i + \textbf{b}_{q} \\
k_i = \textbf{W}_k h_i + \textbf{b}_{k} \\
s_{ij} = q_i^T k_j
\end{array}
\label{eqn:gp}
\end{equation}
where $\textbf{W}_{q,k}$ and $\textbf{b}_{q,k}$ are learnable parameters;
$s_{ij}$ is the predicted score of $(i,j)\in L$. 

\noindent Model weights are optimized by a class-imbalance loss which is also introduced in \citep{su2022global}:
\begin{equation}
\resizebox{\linewidth}{!}{${{\mathcal{L}}} = \log\left(1+\sum\limits_{(i,j)\notin L\atop 1\leq i,j\leq N}{e^{s_{ij}}}\right) + \log\left(1+\sum\limits_{(i,j)\in L\atop 1\leq i,j\leq N}{e^{-s_{ij}}}\right)$}
\end{equation}

\noindent During inference, the model identifies reading order relation pairs by filtering the predictions with a threshold:
\begin{equation}
\hat{L}=\{(i,j)|s_{ij}>0\}
\end{equation}
where $\hat{L}$ is the set of predicted reading order relation pairs.  

\begin{figure}[t]
    \centering
    \vspace{-2mm}
    \includegraphics[width=0.96\linewidth]{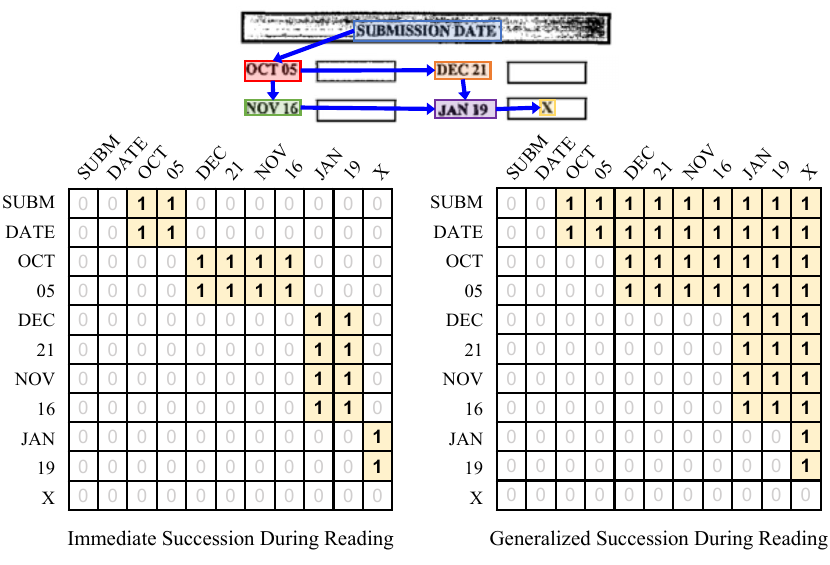}   
    \vspace{-2mm}
    \caption{
        Reading order relation information is represented as a $n*n$ binary matrix to be leveraged in downstream VrD tasks, where $n$ is the number of input textual tokens. 
        }
    \vspace{-4mm}
    \label{fig:ro_attn}
\end{figure}

\subsection{Enhancing Layout-Aware Models with Reading Order Relation}
\label{sec:method_ie}

In this section, it is assumed that the reading order information for a document layout $\mathcal{D}$ is available as a relation between its $N$ layout elements. 
In applications, it could be manually annotated or automatically generated by an off-the-shelf ROP model (see Fig. \ref{fig:rore}). \\
To leverage the reading order information in downstream VrD tasks, we first represent the information in a $n*n$ binary matrix, where $n$ is the number of input textual tokens. 
For each layout element, the pair from each token of itself to each token of its (immediate or generalized) successors is labeled 1 in the matrix, while others are labeled as 0. 
Fig. \ref{fig:ro_attn} displays an example layout whose ISDR and GSDR reading order information are represented as binary matrices.  
The matrix is then used in relation-aware attention modules of downstream layout-aware models. \\
For transformer-based models, the attention matrix within their original self-attention modules is calculated as follows:
\begin{equation}
a_{ij}^{l} = \frac{\exp\left({\textbf{q}_i^l}^T \textbf{k}_j^l / \sqrt{d_k}\right)}{\sum_{j=1}^{n}{\exp\left({\textbf{q}_i^l}^T \textbf{k}_j^l / \sqrt{d_k}\right)}}
\end{equation} 
where 
$\{a_{ij}^l\}_{1\leq i, j\leq n}$ is the attention matrix of a self-attention module in the $l$-th layer; 
$\textbf{q}_i^l, \textbf{k}_i^l$ are the $i$-th query and key vector in the $l$-th layer, respectively; 
$d_k$ is the dimension of the attention head. \\
We propose a relation-aware attention module as a replacement to the original self-attention module, leveraging the reading order relation matrix in the simplest manner. 
This module straightly adds the reading order relation matrix to the original attention matrix to incorporate the information. 
The attention matrix within the relation-aware attention module is calculated as:
\begin{equation}
a_{ij}^{l} = \frac{\exp\left(\left({\textbf{q}_i^l}^T \textbf{k}_j^l + \lambda^l \rho_{ij}\right) / \sqrt{d_k}\right)}{\sum_{j=1}^{n}{\exp\left(\left({\textbf{q}_i^l}^T \textbf{k}_j^l + \lambda^l \rho_{ij}\right) / \sqrt{d_k}\right)}}
\label{eqn:ro_attn}
\end{equation} 
where $\{\rho_{ij}^l\}_{1\leq i, j\leq n}$ is the reading order relation matrix and $\lambda^l$ is a learnable scalar as the weight of reading order information in the $l$-th layer.

\section{Experiments}

\subsection{Experimental Settings}
\label{sec:exp_settings}

\paragraph{Reading Order Prediction as Relation Extraction} 
We test the proposed baseline model with word-level and segment-level reading order prediction tasks of ROOR. 
We take the ISDR annotations as the segment-level labels, and utilize the arranged words within segments together with the segment-level annotations to generate word-level labels. 
We adopt LayoutLMv3-base/large \citep{huang2022layoutlmv3} as the backbone text-and-layout encoder for the proposed ROP model.
In evaluation, we follow the train/validation split from EC-FUNSD and models are evaluated by F1 score. 

\paragraph{Enhancing Downstream VrD Tasks with Reading Order Relation} 
To examine the effectiveness of reading order relation information in enhancing downstream VrD tasks, we enhance pre-trained text-and-layout models (PTLMs) with additional reading order inputs. 
Two baseline models are examined: LayoutLMv3 as the most popular PTLM, and GeoLayoutLM \citep{cvpr2023geolayoutlm} as a representative relation-aware PTLM. 
We assess the baseline models and their reading-order-relation-enhanced (RORE) variants on semantic entity recognition (SER) and entity linking (EL) of EC-FUNSD, leveraging the ground truth reading order information in the ROOR dataset. 

\renewcommand\tabcolsep{5pt}
\begin{table}[t]
\centering
\small
\vspace{-7mm}
\begin{spacing}{1.19}
\centering
\begin{tabular}{l|cc}
    \bottomrule
    \textbf{Method} & \textbf{Word-level} & \textbf{Segment-level} \\
    \hline
    LR \citep{wang2021layoutreader} & - & 9.44 \\
    TPP \citep{zhang2023reading} & - & 42.96 \\
    LayoutLMv3-base & 90.26	& 68.60 \\
    LayoutLMv3-large & \textbf{93.01} & \textbf{82.38} \\
    \hline
    Human & - & 99.28 \\
    \toprule
\end{tabular}
\end{spacing}
\vspace{-2mm}
\caption{The performance of baseline models on reading order relation prediction. Human performance indicates the annotation consistency between two annotators. }
\label{tab:rop_re}
\vspace{-4.5mm}
\end{table}

\paragraph{Quality of Pseudo Reading Order Relation Labels for Downstream Tasks} 
Enhancing an arbitrary VrD task with additional reading order relation inputs is expected. 
However, it is impractical to manually annotate each sample with such information.
Consequently, a robust off-the-shelf ROP model is expected to generalize well to any target domain and generate high-quality pseudo reading order relation labels for unseen datasets, since VrDs in real application could cover various domains and have high diversity in layouts. 

To validate the effectiveness of the generated pseudo labels, we conduct experiments on VrD-IE/QA tasks with utilizing the pseudo reading order information generated by an off-the-shelf LayoutLMv3-large model solely trained on the ROOR dataset.
For IE, we adopt three popular benchmarks for SER and EL: FUNSD \citep{jaume2019funsd}, CORD \citep{park2019cord} and SROIE \citep{huang2019icdar2019}. 
For QA, we choose four popular benchmarks that primarily featuring on layout understanding and IE, but not commonsense or complex reasoning which are less relevant to layout understanding: DocVQA \citep{mathew2021docvqa}, InfoVQA \citep{mathew2022infographicvqa}, WTQ \citep{borchmann2021due} and TextVQA \citep{singh2019towards}. 
A generative variant of LayoutLMv3 with a BART decoder \citep{lewis2019bart} is used as the baseline model, the performance of which is evaluated by average normalized Levenshtein similarity (ANLS). 
We report the performance of baseline and RORE models on these datasets.
The implementation details of the above experiments are illustrated in Appendix \ref{sec:implementation_detail}.

\definecolor{citeblue}{HTML}{000080}
\definecolor{green1}{HTML}{003300}
\definecolor{green2}{HTML}{006600}
\definecolor{green3}{HTML}{009900}
\definecolor{greenmax}{HTML}{00dd00}

\renewcommand\tabcolsep{2.8pt}
\begin{table}[t]
\centering
\small
\vspace{-7mm}
\begin{spacing}{1.19}
\begin{adjustwidth}{-.12in}{-.12in}
\centering
\resizebox{0.96\linewidth}{!}{
\begin{tabular}{c|cc|cc}
    \bottomrule
    \textbf{Method} & \textbf{SER} & \textbf{+RORE} & \textbf{EL} & \textbf{+RORE} \\
    \hline
    LayoutLMv3-base & 82.30 & 82.80$_{\ }$\textcolor{green1}{($\uparrow$0.50)} & 67.47 & 73.64$_{\ }$\textcolor{greenmax}{($\uparrow$6.17)} \\
    LayoutLMv3-large & 83.88* & \textbf{84.53}$_{\ }$\textcolor{green2}{($\uparrow$0.65)} & 78.14* & 79.33$_{\ }$\textcolor{green3}{($\uparrow$1.19)} \\
    GeoLayoutLM & 83.62 & 84.34$_{\ }$\textcolor{green2}{($\uparrow$0.72)} & 86.18 & \textbf{87.42}$_{\ }$\textcolor{green3}{($\uparrow$1.24)} \\
    \toprule
\end{tabular}
}
\end{adjustwidth}
\end{spacing}
\vspace{-2mm}
\caption{Performance of LayoutLMv3 and GeoLayoutLM and their corresponding RORE methods on EC-FUNSD. 
Reproduced results are marked with * (see Appendix \ref{sec:implementation_detail}).}
\label{tab:ec}
\vspace{-4.5mm}
\end{table}

\renewcommand\tabcolsep{5pt}
\begin{table*}[t]
\small
\centering
\vspace{-7mm}
\begin{spacing}{1.19}
\begin{adjustwidth}{-.12in}{-.12in}
\hspace{3.8mm} 
\begin{tabular}{c|l|c|cccc|c}
\bottomrule
\multicolumn{2}{c|}{\multirow{2}{*}{\textbf{Method}}} & \multirow{2}{*}{\textbf{\#Params}} & \multicolumn{2}{c}{\textbf{FUNSD}} & \textbf{CORD} & \textbf{SROIE} & \textbf{Ref. Overall} \\
\multicolumn{2}{c|}{} & & SER & EL & SER & SER & \textbf{Performance} \\
\hline
\multirow{8}{*}{\makecell[c]{Attention}} 
 & SPADE \citep{hwang2021spatial} & base & 72.0 & 41.3 & - & - & 64.95 \\
 & SERA \citep{zhang2021entity} & base & - & 65.96 & - & - & 62.91 \\
 & LiLT \citep{wang2022lilt} & base & 88.41 & 62.76 & 96.07 & - & 85.56 \\
 & DocTr \citep{Liao_2023_ICCV} & 153M & 84.0 & 73.9 & 98.2 & - & 87.87 \\
 & BROS \citep{hong2022bros} & 340M & 84.52 & 77.01 & 97.28 & 96.62 & 88.85 \\
 & LAGaBi \citep{zhu2023beyond} & 133M & 91.00 & - & 97.05 & - & 87.72\\
 & LayoutLMv3 \citep{huang2022layoutlmv3} & 133M & 90.85 & 69.80 & 95.91* & 94.80* & 87.84 \\
 & LayoutLMv3 \citep{huang2022layoutlmv3} & 368M & 91.70* & 79.37* & 96.98* & 96.12* & 91.04 \\
\hline
\multirow{8}{*}{Graph} 
 & Doc2Graph \citep{gemelli2022doc2graph} & base & - & 53.36 & - & - & 50.89 \\
 & RE$^2$ \citep{ramu2023re} & base & - & 71.76 & - & - & 68.44 \\
 & MatchVIE \citep{tang2021matchvie} & base & 81.33 & - & - & 96.57 & 83.17 \\
 & DocFormer \citep{appalaraju2021docformer} & 502M & 84.55 & - & 96.99 & - & 84.96 \\
 & FormNet \citep{lee-etal-2022-formnet} & large & 84.69 & - & 97.28 & - & 85.17 \\
 & mmLayout \citep{wang2022mmlayout} & large & 86.49 & - & 97.38 & 97.91 & 86.34 \\
 & GraphDoc \citep{10.1109/TMM.2022.3214102} & 265M & 87.77 & - & 96.93 & \textbf{98.45} & 86.76 \\
 & DocGraphLM \citep{wang2023docgraphlm} & base & 88.77 & - & 96.93 & - & 86.71 \\
 & GraphLayoutLM \citep{li2023enhancing} & 372M & - & - & 97.98* & - & 86.17 \\
\hline
\multirow{7}{*}{\makecell[c]{Pre-trained}} 
 & StrucTexT \citep{li2021structext} & 107M & 83.09 & 44.1 & - & 96.88 & 79.62 \\
 & DocReL \citep{DocRel-li2022relational} & 142M & - & 46.1 & 97.0 & - & 80.19 \\
 & ERNIE-Layout \citep{peng2022ernie} & large & 93.12 & - & 97.21 & 97.55 & 88.83 \\
 & Bi-VLDoc \citep{luo2022bi} & 409M & 93.44 & - & 97.84 & - & 89.15 \\
 & LayoutMask \citep{tu2023layoutmask} & 404M & 93.20 & - & 97.19 & 97.27 & 88.81 \\
 & Wukong-Reader \citep{bai2023wukong} & 470M & \textbf{93.62} & - & 97.27 & 98.15 & 89.14 \\ 
 & GeoLayoutLM \citep{cvpr2023geolayoutlm} & 399M & 91.10 & 88.06 & 98.11* & 96.62* & 93.47 \\
\hline
\multirow{2}{*}{\makecell[c]{Supervised\\Pre-trained}} 
 & ESP \citep{yang2023modeling} & 50M & 91.12$^\dagger$ & 88.88$^\dagger$ & 95.65$^\dagger$ & - & 92.43 \\
 & UNER \citep{tu2024uner} & base & - & - & - & 97.61$^\dagger$ & 85.99 \\
\hline
\multirow{6}{*}{\textbf{Ours}} 
 & \multirow{4}{*}{RORE-LayoutLMv3} & \multirow{2}{*}{133M\textcolor{blue}{+12}} & 91.39 & 71.69 & 96.72 & 95.65 & 88.86 \\
 &  &  & \textcolor{green2}{($\uparrow$0.54)} & \textcolor{green3}{($\uparrow$1.89)} & \textcolor{green2}{($\uparrow$0.81)} & \textcolor{green2}{($\uparrow$0.85)} & \textcolor{green3}{($\uparrow$1.02)} \\
 \cline{3-8}
 & & \multirow{2}{*}{368M\textcolor{blue}{+24}} & 92.14 & 80.84 & 97.29 & 96.59 & 91.71 \\
 & & & \textcolor{green1}{($\uparrow$0.44)} & \textcolor{green3}{($\uparrow$1.47)} & \textcolor{green1}{($\uparrow$0.31)} & \textcolor{green1}{($\uparrow$0.47)} & \textcolor{green2}{($\uparrow$0.67)} \\
 \cline{2-8}
 & \multirow{2}{*}{RORE-GeoLayoutLM} & \multirow{2}{*}{399M\textcolor{blue}{+24}} & 91.84 & \textbf{88.46} & \textbf{98.52} & 96.97 & \textbf{93.94} \\
 & & & \textcolor{green2}{($\uparrow$0.74)} & \textcolor{green1}{($\uparrow$0.40)} & \textcolor{green1}{($\uparrow$0.41)} & \textcolor{green1}{($\uparrow$0.35)} & \textcolor{green1}{($\uparrow$0.47)} \\
\toprule
\end{tabular}
\end{adjustwidth}
\end{spacing}
\vspace{-2mm}
\caption{
Performance of prevailing methods on three VrD-IE benchmarks.  
Best results are marked bold. 
Reproduced results are marked with * (see Appendix \ref{sec:implementation_detail}).
Supervised pre-trained results (marked with $\dagger$) are not directly comparable to our methods and are listed as reference (see Appendix \ref{sec:baseline_selection}).
"Base" and "large" models have 100\textasciitilde 200M and 300\textasciitilde 400M parameters, respectively. 
Referential overall performance is calculated by averaging on the imputed score matrix using nuclear norm minimization. 
}
\vspace{-3mm}
\label{tab:main}
\end{table*}

\subsection{Performance of Baseline ROP Models}
\label{sec:exp_ro}

We compare the proposed relation-based baseline models with previous sequence-based models which predict a layout element sequence as the reading order \citep{wang2021layoutreader,zhang2023reading}.
We report their performance in Tab. \ref{tab:rop_re} together with human performance.
Without task-specific design, the relation-based models are capable to learn the reading order relation signals, which reveals their potential to generate pseudo reading order annotations for other datasets. 
However, the sequence-based models struggle with incomplete supervision and demonstrate a significant performance gap. 
We have also found that: (1) The proposed model design benefits from stronger pre-trained backbones, as the performance of the LayoutLMv3-large model is much better than the base model in Tab. \ref{tab:rop_re}. 
(2) The proposed models generally perform better when the task level and the granularity of bounding box inputs coincide, illustrated by an ablation study in Appendix \ref{sec:ablation}.
(3) The proposed models are good at identifying non-linear cases, yet showing the tendency to over-rely on layout but not textual features, demonstrated by visualized case studies in Appendix \ref{sec:case}. 

\renewcommand\tabcolsep{2.8pt}
\begin{table}[t]
\centering
\small
\begin{spacing}{1.19}
\begin{adjustwidth}{-.12in}{-.12in}
\centering
\begin{tabular}{c|cc|cc}
    \bottomrule
    \textbf{Task} & \textbf{DocVQA} & \textbf{+RORE} & \textbf{InfoVQA} & \textbf{+RORE} \\
    \hline
    \textbf{Score} & 69.13 & 73.53$_{\ }$\textcolor{greenmax}{($\uparrow$4.40)} & 22.62 & 29.20$_{\ }$\textcolor{greenmax}{($\uparrow$6.48)} \\
    \specialrule{.1em}{.05em}{.05em} 
    \textbf{Task} & \textbf{WTQ} & \textbf{+RORE} & \textbf{TextVQA} & \textbf{+RORE} \\
    \hline
    \textbf{Score} & 39.94 & 41.71$_{\ }$\textcolor{green3}{($\uparrow$1.77)} & 45.15 & 45.24$_{\ }$\textcolor{green1}{($\uparrow$0.09)} \\
    \toprule
\end{tabular}
\end{adjustwidth}
\end{spacing}
\vspace{-2mm}
\caption{Performance of the baseline generative method and RORE variant on four VrD-QA benchmarks.} 
\label{tab:docvqa}
\vspace{-6mm}
\end{table}

\subsection{Reading Order Relation Enhanced Methods for Downstream VrD Tasks}
\label{sec:exp_ie}

As illustrated in \S\ref{sec:method_ie}, we incorporate the reading order relation information to baseline models with the relation-aware attention module introduced in Eqn. \ref{eqn:ro_attn}. 
In Tab. \ref{tab:ec}, \ref{tab:main} and \ref{tab:docvqa}, we compare the performance of RORE methods on VrD-IE/QA tasks with their corresponding baseline methods, together with other competitive methods of different categories. 

We conclude three general trends from the experiment results, demonstrating the effectiveness of the proposed methods: 
(1) \textbf{The performance of downstream tasks are universally improved by proper leveraging the reading order relation information.} 
As shown in Tab. \ref{tab:ec}, the performance of three RORE methods have shown remarkable improvements and achieved SOTA on both two tasks. 
Significantly, these models benefit from a +2.86 average performance improvement on EL. 
(2) \textbf{Our ROP model is able to generalize to unseen datasets. }
Tab. \ref{tab:main} displays that all of the four task settings over three baseline methods have substantially improved by RORE, with only leveraging the pseudo labels generated by an off-the-shelf ROP model.
Tab. \ref{tab:docvqa} demonstrates a similar trend on generative VrD-QA tasks. 
The results demonstrate that our pipeline with an off-the-shelf ROP model trained on ROOR is capable to generate usable reading order labels for any arbitrary VrD tasks, without any additional manual annotations. 
(3) \textbf{The proposed RORE methods are very competitive among various tasks.} 
Although we do not pursue exceedingly good performance on cross-domain settings, the RORE methods achieve SOTA on two of them. Especially, RORE-GeoLayoutLM achieves the best comprehensive performance among all prevailing methods on VrD-IE in Tab. \ref{tab:main}. 
Besides, the feasibility of using pseudo reading order labels is further verified in Appendix \ref{sec:ablation}.

Other notable findings according to the experiment results include:
(1) EL generally benefits more than SER by leveraging the reading order relation information, according to Tab. \ref{tab:ec} and \ref{tab:main}. 
Compared with SER which focuses on capturing entity semantics, EL relies more on the modeling of logical structure and spacial relationships of the layout, which is fully modeled by the reading order relation information. 
(2) The enhancement brought by RORE is particularly critical in low-resource scenarios. 
Among the baseline methods, LayoutLMv3-base initially exhibits the worst performance, yet it achieves the greatest improvement after incorporating the reading order relation information, revealing the potential of our method in low-resource scenarios. 
(3) The performance improvement through RORE should be primarily attributed to the auxiliary input information, rather than the relation-aware architecture. 
Among the VrD-QA benchmarks, TextVQA is unique from the others with its relatively simple layout patterns. Most samples in TextVQA are book covers, within whose layouts the text regions are scarce and generally adhere to a from-top-to-bottom reading order, thus the reading order relation inputs are less informative.
According to Tab. \ref{tab:docvqa}, TextVQA benefits less from RORE than the other benchmarks, from which we conclude that RORE primarily benefits from the additional inputs but not the architecture. 
(4) ISDR and GSDR benefits the downstream tasks to a similar extent, discussed by an ablation study in Appendix \ref{sec:ablation}.
Summarizing all the above points, we demonstrate our final conclusion: \textbf{the reading order relation information is an informative auxiliary feature, from which RORE methods learn reliable and robust human reading patterns, attributing to their effectiveness.} The conclusion is further verified by a case study in Appendix \ref{sec:case}. 

\section{Related Work}
\label{sec:related_work}

\paragraph{Modeling of Layout Reading Order}
The assumption that layout reading order could be represented as a permutation on layout elements can be traced back to \citep{nagy1984hierarchical}. 
Since then, rule-based methods \citep{ishitani2003document,ferilli2014abstract}, statistical-ML-based methods \citep{han2007frame,malerba2007learning} and deep-learning-based methods \citep{li2020end,zhang2023reading} have been proposed.
However, several previous works have noted that layout reading order information cannot be fully conveyed by one single permutation of layout elements. These works improved their utilization of reading order information according to specific application requirements \citep{gu2022xylayoutlm,tu2023layoutmask}. 
However, these works do not delve further into the underlying problem that the formulation of layout reading order should be improved.  

In this work, we model the reading order as ordering relations over layout elements, inspired by previous works which modeled the non-linear structures of VrD layouts as binary relations \citep{aiello2003bidimensional, aiello2004thick} or ordered structures  \citep{clausner2013significance}, and also order-theoretic modeling of linguistic objects \citep{vendrov2015order,athiwaratkun2018hierarchical,liu-etal-2023-linear}. 
A more comprehensive introduction of the evolution of modeling and utilizing layout reading order is available at Appendix \ref{sec:related_work_rop}.

\paragraph{Modeling Relation between Layout Elements}
Several layout-aware models have devised relation-aware modules for better modeling VrD layouts. 
\citet{zhao2022read,pham2022understanding} enhanced local awareness of layout tokens with specialized attention mechanisms, and \citet{chen2023global} introduced a special module for better modeling the global relationships. 
More works attempted to better leverage the relative positional relationship between layout elements by improving attention mechanism \citep{xu2021layoutlmv2,xu2021layoutxlm,powalski2021going,hwang2021spatial,nguyen2021skim,zhang2021entity,hong2022bros,zhu2023beyond,siyuan-etal-2024-layoutpointer}, utilizing graph structure \citep{liu-etal-2019-graph,luo2020merge,carbonell2021named,gemelli2022doc2graph,luo2022doc,mohbat-etal-2023-gvdoc,ramu2023re,krishnan2024towards}, or relation-aware pre-training \citep{DocRel-li2022relational,wang2022towards,peng2022ernie,cvpr2023geolayoutlm,yu2023structextv}.

\section{Conclusion}

In this paper, we investigate reading order prediction for visually-rich document layouts, a classic and important task of document AI. 
We question the traditional formulation of this task as it is unable to convey the complete reading order information within the layouts, indicating that the improper formulation would possibly affect model performance on downstream VrD tasks.
To address the problem, we propose an improved formulation of this task by conceptualizing layout reading order as ordering relations.
We further prove the effectiveness of utilizing the improved form of reading order information in downstream tasks by extensive experiments.
The proposed reading-order-relation-enhanced methods achieve universal performance improvement on multiple task settings of VrD-IE and VrD-QA, including SOTA performance on two targeted settings and two cross-domain settings. 
To facilitate the development of automated methods on ROP in the improved form, we also establish a benchmark dataset with high-quality reading order relation annotations together with a baseline model, which builds a pipeline to generate usable pseudo reading order labels for any arbitrary downstream tasks.

\section*{Limitations}

Our work has the following limitations: 

\begin{enumerate}[leftmargin=*,noitemsep,topsep=0pt]
    \item There is still room for improving the design of the reading order relation prediction model. In this paper, we propose a baseline relation extraction model which is constituted by an encoding pre-trained text-and-layout representation model and a decoding global pointer network, achieving fairly good result on the reformulated ROP. However, the optimal solution of this problem is still under explored. Appendix \ref{sec:case} have studied the strengths and weaknesses of the proposed model, which we hope would shed light on the future works attempting to improve the model design. 
    \item Several methods for improving VrD understanding with reading order relation information remain insufficiently explored. 
    In \S\ref{sec:related_work}, we have categorized the relation-aware VrD understanding methods to attention-based, graph-based and relation-aware pre-training-based methods. 
    In this paper, we have attempted to fuse the reading order information via a relation-aware attention module, demonstrating that our method improves the performance on downstream tasks
    However, other ways to leverage the reading order information are not attempted and compared with the proposed method. 
    It is encouraged for future works to make attempts on efficiently leverage the reading order relation information with ROOR and the proposed experiment settings. 
\end{enumerate}

\noindent Besides, it is necessary to point out that the improved form of ROP has increased the complexity of this task, as the search space of valid answers has increased from $O(n!)$ to $O(2^{n^2})$ by introducing the reading order relation, where $n$ denotes the number of layout elements. 
We regard the complexity increase as an essential budget for the precise modeling of layout reading order. 

\section*{Acknowledgements}

The authors wish to thank the anonymous reviewers for their helpful comments. This work was partially funded by National Natural Science Foundation of China (No.62376061,62206057), Shanghai Rising-Star Program (23QA1400200), Natural Science Foundation of Shanghai (23ZR1403500), Program of Shanghai Academic Research Leader under grant 22XD1401100, and Ant Group Research Intern Program.

\bibliography{custom}

\newpage

\appendix

\section{Theoretical Foundations of Directed Acyclic Relation and Order Relations}
\label{sec:theory}

To formalize the problem we intend to solve, necessary concepts and fundamental theorems from order theory are introduced in below, where $A$ denotes an arbitrary set: \\
\textbf{Definition 1} (Directed Acyclic Relation)\textbf{.}\\
A binary relation $DAR \in A \times A$ is a directed acyclic relation over $A$ iff. it is: 
\begin{enumerate}[leftmargin=*,noitemsep,topsep=0pt]
    \item Acyclic: $\forall k \geq 1 : \nexists \{s_i\}_{1 \leq i \leq k} \subseteq A$ s.t. $\{ (s_i, s_{i+1}) \}_{1 \leq i \leq k} \subseteq DAR$ where $s_{k+1} = s_1$.
\end{enumerate}
\textbf{Comment 1.} Each directed acyclic relation $R$ over $A$ corresponds to a directed acyclic graph $G = (A, E)$, where the nodes are elements in $A$, and each edge $(s_1, s_2) \in E$ is a linked node pairs in $R$ such that $s_1 R s_2$. \\
\textbf{Definition 2} (Partial Order, Strict Partial Order)\textbf{.}\\
A binary relation $PO \in A \times A$ is a partial order over $A$ iff. it is: 
\begin{enumerate}[leftmargin=*,noitemsep,topsep=0pt]
    \item Reflexive: $\forall s \in A : (s, s) \in PO$.
    \item Antisymmetric: $\forall s_1, s_2 \in A: (s_1, s_2) \in PO \land (s_2, s_1) \in PO \Rightarrow s_1 = s_2$.
    \item Transitive: $\forall s_1, s_2, s_3 \in A: (s_1, s_2) \in PO \land (s_2, s_3) \in PO \Rightarrow (s_1, s_3) \in PO$.
\end{enumerate}
A binary relation $SPO \in A \times A$ is a strict partial order over $A$ iff. it is antisymmetric, transitive and: 
\begin{enumerate}[leftmargin=*,noitemsep,topsep=0pt]
    \item Irreflexive: $\forall s \in A : (s, s) \notin SPO$.
\end{enumerate}
\textbf{Definition 3} (Transitive Closure)\textbf{.}\\
The transitive closure of a binary relation $R$ over $A$ is the smallest (w.r.t. $\subseteq$) transitive relation $R^+$ over $A$ such that $R \subseteq R^+$.\\
\textbf{Theorem 1.} For any relation $R$ over $A$, the transitive closure of $R$ always uniquely exists. \\
\textbf{Proof of Theorem 1.} The proof of existence and uniqueness are required of the transitive closure of an arbitrary binary relation $R$ over $A$. 
\begin{enumerate}[leftmargin=*,noitemsep,topsep=0pt]
    \item Proof of existence: Note that the intersection of any family of transitive relations is again transitive. For an arbitrary binary relation $R$ over $A$, there exists at least one transitive relation containing $R$, namely the trivial one: $A\times A$. The transitive closure of $R$ is then given by the intersection of all transitive relations containing $R$ as $R^+=\underset{R'\subseteq A\times A, R\subseteq R'}{\bigcap}R'$. \\It is trivial to see that $R^+$ is transitive and is the smallest among all transitive relations containing $R$. 
    \item Proof of uniqueness: Given that there exist $R^1$ and $R^2$ and both of them are the transitive closures of $R$, note that $R^2$ is a transitive relation over $A$, and therefore $R^1 \subseteq R^2$. Similarly $R^2 \subseteq R^1$, therefore $R^1 = R^2$. 
\end{enumerate} 
\textbf{Theorem 2.} The transitive closure of a DAR is a SPO. \\
The proof of Theorem 2 is based on Lemma 1. \\
\textbf{Lemma 1. } For the transitive closure $R^+$ of an arbitrary DAR $R$ over set $A$, there do not exist $s_1, s_2 \in A$ such that $(s_1, s_2) \in R^+$ and $(s_2, s_1) \in R^+$. \\
\textbf{Proof of Lemma 1. } Assume that $(s_1, s_2) \in R^+$ and $(s_2, s_1) \in R^+$. By the definition of the transitive closure, there exist sequences of elements $(a_1, a_2, \dots, a_n), \{a_i\}_{1\leq i\leq n}\subseteq A$ and $(b_1, b_2, \dots, b_m), \{b_j\}_{1\leq j\leq m}\subseteq A$ such that: \\
$(a_i, a_{i+1}) \in R$ for all $1 \leq i < n$, where $s_1 = a_1, a_n = s_2$; $(b_j, b_{j+1}) \in R$ for all $1 \leq j < m$, where $s_2 = b_1, b_m = s_1$. \\
However, $(a_1, a_2, \dots, a_n=b_1, b_2, \dots, b_m=a_1)$ comprises a cycle within $R$, which is conflict with the definition of directed acyclic relation. 
Therefore, the assumption does not hold, proving that there do not exist $s_1, s_2 \in A$ such that $(s_1, s_2) \in R^+$ and $(s_2, s_1) \in R^+$. \\
\textbf{Proof of Theorem 2.} The transitive closure $R^+$ of an arbitrary DAR $R$ over set $A$ is:
\begin{enumerate}[leftmargin=*,noitemsep,topsep=0pt]
    \item Irreflexive: When it takes $s=s_1=s_2$ in Lemma 1, $\forall s \in A, (s, s) \notin R^+$.
    \item Antisymmetric: Due to Lemma 1, the premise such that $(s_1, s_2) \in R^+ \land (s_2, s_1) \in R^+$ is always false for all $s_1, s_2 \in A$, so this property is always valid.
    \item Transitive: Since $R^+$ is a transitive closure, this property is always valid.
\end{enumerate}
Therefore, $R^+$ is a SPO over $A$ by definition. \\
\textbf{Definition 4} (Total Order, Strict Total Order)\textbf{.}\\
A partial order $TO$ over $A$ is a total order iff. $\forall s_1, s_2 \in A: (s_1, s_2) \in TO \lor (s_2, s_1) \in TO$, i.e. every two elements of $A$ are comparable.  \\
A strict partial order $STO$ over $A$ is a strict total order iff. $\forall s_1, s_2 \in A \land s_1 \neq s_2: (s_1, s_2) \in STO \lor (s_2, s_1) \in STO$, i.e. every two different elements of $A$ are comparable. \\
\textbf{Comment 2.} Each TO/STO over $A=\{s_i\}_{1\leq i\leq |A|}$ can be represented as a permutation of $A$ as $(s_{P(1)},\dots,s_{P(|A|)})$ where $P$ is a bijection from $\{i\}_{1\leq i\leq |A|}$ to itself. 

\section{A Brief Overview of Layout Reading Order Prediction}
\label{sec:related_work_rop}

Generally, the term layout reading order refers to the order in which people will read layout contents. 
Modeling the order itself plays an important role to visually-rich document understanding, while it is also beneficial to incorporate the reading order information in addressing various VrD tasks. 
Also, the cognition of this problem is advancing along with the development of this field.
This section introduces the progressive development of modeling of layout reading order, together with its application in document intelligence. 

\paragraph{Reading Order as a Permutation}
The assumption that layout reading order could be represented as a permutation on layout elements can be traced back to \citep{nagy1984hierarchical}, which proposed XY-cut tree to represent the hierarchical structure of layout elements. 
A permutation of layout elements can be derived from the XY-cut tree as the layout reading order. 
Rule-based methods \citep{ha1995recursive,ishitani2003document,meunier2005optimized} and statistical-ML-based methods \citep{han2007frame,7351614} have been proposed to predict the optimal XY-cut tree of layouts. 
However, the layout reading order is not always from-top-to-bottom and from-left-to-right, which constrains the maximum effectiveness of XY-cut tree in representing the reading order. 
Therefore, \citet{malerba2007learning,malerba2008machine,ferilli2014abstract} have proposed to directly predict a permutation of layout elements, rather than a XY-cut tree. 

Coming to the deep learning era, \citet{li2020end} proposed a pointer network with a GCN layout encoder for reading order prediction. 
Following this paradigm, ORO \citep{9413256,quiros2022reading}, DocReL \citep{DocRel-li2022relational}, TPP \citep{zhang2023reading} and MLARP \citep{QIAO2024110314} tackled the task via devising relation extraction models between layout elements and decoding strategies from predicted relation pairs. 
Besides, LayoutReader \citep{wang2021layoutreader} predicted a word sequence as the reading order sequence without guaranteeing its property as a permutation of input words. 
SGS \citep{wang2023text} introduced two major patterns of reading order as column-wise and row-wise order, and tackled the task by segmenting and sorting ordered blocks. 

In leveraging reading order information in downstream application, ERNIE-Layout \citep{peng2022ernie} includes a pre-training objective of reading order prediction; 
TPP \citep{zhang2023reading} verifies the effectiveness of a reading order prediction model to correct the input token sequence of layout-aware models. 
Relevant data resources that are publicly available include ReadingBank \citep{wang2021layoutreader} and DocTrack \citep{wang2023doctrack} on scanned VrDs, and OHG, FCR and ABP \citep{quiros2022reading} on handwritten VrDs. 
However, all the above works failed to capture or leverage the complete reading order information due to the inappropriate modeling of reading order. 

\paragraph{Reading Order beyond a Permutation}
Several previous works have noted that layout reading order information cannot be fully conveyed by one single permutation of layout elements, i.e. a globally total-ordered element sequence. 
All these works treated the incomplete reading order information issue as a task-irrelevant inductive bias injecting noise into the inputs, and attempted to mitigate its negative effects.
In transformer-based layout-aware models, the noise brought by noisy (\textit{incomplete}) reading order information is usually introduced by absolute position embeddings. 
LSPE \citep{wang2022simple} introduced a learnable fully-connected feed-forward sinusoidal positional embedding network to handle the noise. 
XYLayoutLM \citep{gu2022xylayoutlm} reduced the random noise by utilizing multiple globally total-ordered element sequences generated by an augmented XY-cut algorithm. 
UTel \citep{tao2022knowing} introduced 1D clipped relative position embeddings to mitigate the use of long-term reading order information which is much more noisy. 
LayoutMask \citep{tu2023layoutmask} utilized locally total-ordered word sequences of segments with local 1D position embeddings. 
These works increased the robustness of models on the noisy reading order information, but did not make use of the non-linear reading order information which is essential in complex layouts.
From theoretical perspective, \citet{ceci2007data} modeled the layout reading order as multiple disjoint chains (i.e. total-ordered subsets) of the set of layout elements. 
However, this assumption did not consider the numerous intertwined reading order relationships in irregular layouts, such as design layouts (Fig. \ref{fig:motivation}a) and tables (Fig. \ref{fig:motivation}c).

\paragraph{Reading Order as Ordering Relations}
In this work, we model the reading order as ordering relations over layout elements. 
The presented modeling draws inspiration from previous works which modeled other non-linear structures of VrD layouts as binary relations or ordered structures.
\citet{aiello2003bidimensional, aiello2004thick} illustrated to represent the reasoning dependencies about temporal intervals between layout elements as a bi-dimensional Allen relation \citep{allen1983maintaining}. 
\citet{clausner2013significance,shrivastava2021handling} introduced tree structures of ordered and unordered groups to represent the logical structure of layout elements using a tree of ordered and unordered groups. 

The presented modeling is also motivated by order-theoretic modeling of linguistic objects.
\citet{vendrov2015order} proposed to model the visual-semantic hierarchy as a partial order  over images and language, as it can be displayed as a DAG.
\citet{athiwaratkun2018hierarchical} introduced density order embeddings to embed the hierarchy relation into vector space.
\citet{liu-etal-2023-linear} presented an order-theoretic framework to model the relation between token pairs, which universally tackled a series of structured prediction tasks.
These works highlights the motivation of predicting and leveraging the inherent structures within textual data by modeling them as ordering structures. 

\section{Baseline Selection for Semantic Entity Recognition and Entity Linking}
\label{sec:baseline_selection}

Semantic entity recognition and entity linking are the most typical and important tasks in VrD-IE. 
On tackling these tasks, task-specific methods are developed together with universal methods suited to multiple tasks. 
Also, the performance on these tasks are adopted as an evaluation of the performance of pre-trained text-and-layout models. 
In order to conduct a comprehensive comparison between our methods and other diverse methods, we roughly categorize the existing methods, selecting the most representative methods from each category for comparison. 
In principle, we only consider layout-based fully-supervised methods, and mainly focus on the most popular methods from the collection of top-tier venues (conferences and journals) in AI applications:
\begin{itemize}[leftmargin=*,noitemsep,topsep=0pt]
    \item \textbf{Attention}: These methods attempt to better capture task-relevant features by modulating the original attention mechanism. 
    Especially, entity linking methods propose relation-aware attention modules to model the linking relationship between layout elements. 
    We select representative text-and-layout representation learning methods \citep{wang2022lilt,hong2022bros,huang2022layoutlmv3,zhu2023beyond} and entity linking methods \citep{hwang2021spatial,zhang2021entity,Liao_2023_ICCV} for comparison.
    \item \textbf{Graph}: These methods tackle the task by modeling graph structures, which is especially useful in modeling the linking relationships between entities in EL task. 
    In representative methods, GNN-based structures are adopted for either boosting relation-aware pre-training \citep{appalaraju2021docformer,lee-etal-2022-formnet,wang2022mmlayout,10.1109/TMM.2022.3214102,wang2023docgraphlm,li2023enhancing} or entity linking modeling \citep{tang2021matchvie,gemelli2022doc2graph,ramu2023re}.
    \item \textbf{Pre-trained}: These methods take advantage of unsupervised pre-training on large-scale data resources. They achieve the goal only by designing pre-training objectives, rather than introducing specific architectures. 
    The selected methods are all pre-trained text-and-layout models aiming to serve as a representation model for document layouts \citep{li2021structext,DocRel-li2022relational,peng2022ernie,luo2022bi,tu2023layoutmask,bai2023wukong,cvpr2023geolayoutlm}.
    \item \textbf{Supervised pre-trained}: These methods leverage extra supervised data of the downstream tasks, therefore their performance are not fairly comparable to our methods \citep{yang2023modeling, tu2024uner}. 
    Nevertheless, considering the similar research aim and model paradigm, the results of these methods are also reported in the main table. 
    \item \textbf{LLM-based}: These methods are based on large-scale generative pre-training or supervised fine-tuning on domain-specific data resources. 
    Considering their text-generation paradigm and disproportionate parameter scale, the application scenario of these models should be totally different from the extraction-based methods introduced above. 
    Therefore, we separately listed their performance in Tab. \ref{tab:llm} to compare with our methods.  
    \item \textbf{Span-extraction-based}: These methods are also worth mentioned as they have also achieved good performance on VrD-IE tasks \citep{zhang2023reading,li2024hypergraph}. 
    However, due to the difference on the evaluation metrics, these methods are not involved in the comparison. 
\end{itemize}

\section{Implementation Details}
\label{sec:implementation_detail}

\paragraph{Reading Order Prediction Methods} 

For training the baseline sequence-based ROP methods (LR and TPP) on ROOR, these models are supervised by the topology-sorted sequence of the original label for each sample. 
The training process generally follows the settings reported in the original papers, with the only exception that we fine-tune each model by 500 epoches. 
During inference, these models produce word sequences as outputs, and each two adjacent words are extracted as a predicted reading order relation pair. 
In segment-level evaluation of these models, we order the segments according to the first appearance of their words in the original outputs, constructing segment sequences as outputs. 

\renewcommand\tabcolsep{2.8pt}
\begin{table}[t]
\centering
\small
\begin{spacing}{1.19}
\begin{adjustwidth}{-.12in}{-.12in}
\centering
\resizebox{0.95\linewidth}{!}{
\begin{tabular}{c|c|ccc}
    \bottomrule
    \textbf{Method} & \textbf{\#Params} & \textbf{FUNSD} & \textbf{CORD} & \textbf{SROIE} \\
    \hline
    UDOP \citeyearpar{tang2023unifying} & 794M & 91.62 & 97.58 & - \\
    DocLLM \citeyearpar{wang2023docllm} & 7B & 51.8 & 67.4 & 91.9 \\
    LayoutLLM \citeyearpar{luo2024layoutllm} & 7B & 79.98 & 63.10 & 72.12 \\
    \hline
    RORE-GeoLayoutLM & \textasciitilde399M & \textbf{91.84} & \textbf{98.52} & \textbf{96.97} \\
    \toprule
\end{tabular}
}
\end{adjustwidth}
\end{spacing}
\vspace{-2mm}
\caption{Performance of LLM-based generative methods and our methods on three VrD-SER benchmarks.} 
\label{tab:llm}
\vspace{-6mm}
\end{table}

The proposed relation-extraction-based ROP model adopts LayoutLMv3-base \citep{huang2022layoutlmv3} and LayoutMask \citep{tu2023layoutmask} as the backbones. 
The maximum sequence length of textual tokens for both of them is 2048.
The maximum layout elements is set to 512 for word-level task and 256 for segment-level task, ensuring the number of layout elements within each ROOR samples should not exceeds the maximum.
The head size of the global pointer decoding network (i.e. the dimension of query and key vectors in Eqn. \ref{eqn:gp}) is set to 128. 
We use an AdamW optimizer with 5\% linear warming-up steps, a 0.1 dropout rate, and a 1e-5 weight decay, together with a cosine scheduler. 
The learning rate is searched from \{1e-5, 2e-5, 5e-5, 1e-4, 1.5e-4, 2e-4\} for LayoutLMv3-base and \{1e-5, 2e-5, 5e-5, 1e-4\} for LayoutLMv3-large, and the best learning rate is 1e-4/2e-5, respectively. 
We fine-tune the model by 1,000 epoches (10,000 steps) on 1 A800 GPUs with batch size being 16 and patience being 50 for early stopping. 

\paragraph{Reading Order Relation Enhanced Methods for Downstream VrD Tasks}
For information extraction tasks, the enhanced semantic
entity recognition (SER) and entity linking (EL) models strictly keep the same settings of the original implementation \citep{zhang2024rethinking}, with all experiments using the learning rate of 1e-5, batch size of 16, number of fine-tuning epoches of 500 and early stopping patience of 50. 
The initial value for $\lambda^l$ is all set to 10 in all layers, and the learning rate of the $\lambda$s are searched between the original learning rate and 1e-2, where the latter is for faster convergence. 
Specially, in these three groups of experiment: LayoutLMv3-base-FUNSD-SER, GeoLayoutLM-FUNSD-SER and GeoLayoutLM-FUNSD-EL, since we used the fine-tuned weights provided by the official repositories of LayoutLMv3 \footnote{\url{https://github.com/microsoft/unilm/tree/master/layoutlmv3}} and GeoLayoutLM \footnote{\url{https://github.com/AlibabaResearch/AdvancedLiterateMachinery/tree/main/DocumentUnderstanding/GeoLayoutLM}} which are fairly close to the optimal state, we only use the reading order information in the first 4 layers, and the value for $\lambda^l, 1 \leq l \leq 4$ are set to 0.1 initially. 
In Tab. \ref{tab:ec} and \ref{tab:main}, for the baseline models, we report the bias-removed running results from \citep{zhang2024rethinking} together with our running results using its implementation.
Among other baseline methods, the performance of GraphLayoutLM \citep{li2023enhancing} can only be reproduced on CORD according to its official release \footnote{\url{https://github.com/Line-Kite/GraphLayoutLM}}. We are unable to reproduce its performance on other datasets since the pre-trained weights are not released. 
Therefore, we only report its performance on CORD from its official release.
These reproduced results are marked with *.

For generative VrD-QA tasks, the baseline model is comprised of a LayoutLMv3-base encoder and a BART-base decoder \citep{lewis2019bart}, and is pre-trained on open-sourced VrD-QA-form data filtered from InstructDoc \citep{tanaka2024instructdoc} for better initialization. 
The model is then fine-tuned on each dataset using learning rate of 1e-5, batch size of 48, with an AdamW optimizer with 2\% linear warming-up steps and a cosine scheduler for 6 epochs. 
For the RORE settings, the reading-order-relation-aware attention module is only applied to its decoder, and the other settings keep the same of RORE-VrD-IE experiments. 

\renewcommand\tabcolsep{5pt}
\begin{table}[t]
\centering
\small
\begin{spacing}{1.19}
\centering
\begin{tabular}{c|c|cc}
    \bottomrule
    \multirow{2}{*}{\makecell[c]{\textbf{Task Level}}} & \multirow{2}{*}{\makecell[c]{\textbf{Method}}} & \multicolumn{2}{c}{\textbf{Bounding Box Level}} \\
    \cline{3-4}
    & & \ \ Word\ \ \ \  & Segment \\
    \hline
    \multirow{2}{*}{\makecell[c]{Word}} 
    & LayoutLMv3-base & \textbf{90.26} & 87.93 \\
    & LayoutLMv3-large & \textbf{93.01} & 92.80 \\
    \hline
    \multirow{2}{*}{\makecell[c]{Segment}} 
    & LayoutLMv3-base & 63.00 & \textbf{68.60} \\
    & LayoutLMv3-large & 76.04 & \textbf{82.38} \\
    \toprule
\end{tabular}
\end{spacing}
\vspace{-2mm}
\caption{Ablation study on the level of bounding box information used in the baseline ROP model.}
\label{tab:rop_re_box_level}
\vspace{-4.5mm}
\end{table}

\section{Ablation Studies}
\label{sec:ablation}

\paragraph{Impact of Layout-level on ROP Models}

For better understanding the usage of text-and-layout information in the proposed ROP models, we conduct an ablation study by configuring the level of bounding box inputs. 
The results in Tab. \ref{tab:rop_re_box_level} demonstrate that the proposed ROP models generally perform better when the bounding box level coincides with the task level. 

\renewcommand\tabcolsep{5pt}
\begin{table}[t]
\small
\centering
\begin{spacing}{1.19}
\resizebox{1.01\linewidth}{!}{
\begin{tabular}{c|ccc}
\bottomrule
\diagbox{\textbf{Task}}{\textbf{RORE}} & \textbf{None} & \textbf{Pseudo} & \textbf{Ground Truth} \\
\hline
SER & 83.62	& 83.96 \textcolor{green1}{($\uparrow$0.34)} & \textbf{84.34} \textcolor{green2}{($\uparrow$0.72)} \\
EL & 86.18 & 86.56 \textcolor{green1}{($\uparrow$0.38)} & \textbf{87.42} \textcolor{green3}{($\uparrow$1.24)}
\\ 
\toprule
\end{tabular}
}
\end{spacing}
\vspace{-2mm}
\caption{
Ablation study on the impact of the quality of reading order relation supervision. 
We report the performance of GeoLayoutLM and its RORE variant on EC-FUNSD with pseudo or ground truth reading order relation inputs are reported. 
The pseudo labels are generated by the same ROP model used in Tab. \ref{tab:main} and \ref{tab:docvqa}. 
}
\vspace{-4mm}
\label{tab:pseudo_or_gt}
\end{table}

\paragraph{Impact of the Quality of Pseudo Reading Order Relation Labels for Downstream Tasks}

In the introduced RORE pipeline, pseudo reading order relation labels generated by off-the-shelf ROP models are utilized for any VrD task lacking manual reading order annotations. 
It is of interest to determine the extent to which performance is affected when using pseudo labels instead of ground truth labels in downstream tasks. 

To quantify the adverse effects of employing pseudo labels, we conduct an ablation study based on the EC-FUNSD dataset, which has available ground truth reading order labels (from ROOR). We conduct experiments on GeoLayoutLM for VrD-IE, choosing either pseudo or ground truth labels during training and inference.
According to Tab. \ref{tab:pseudo_or_gt}, there is a noticeable performance gap when using pseudo labels. 
These results underscore the importance of creating corresponding reading order relation annotations when implementing the RORE pipeline in real-world VrD applications. 
Additionally, these results reflect the effectiveness of RORE methods in Tab. \ref{tab:main} and \ref{tab:docvqa}, as they have achieved satisfying performance even when the ground truth labels are unavailable and the target task differs from the domain of the ROP model.

\paragraph{Impact of Different Reading Order Relations on Application Performance}

In this work, two types of reading order relationships are proposed, of which ISDR is believed to represent local adjacent relationship between layout element, and GSDR is known as a global order representation indicating the possible order relationship between each two layout elements. 
To determine the role of each of them in enhancing downstream VrD tasks, we conduct experiments on GeoLayoutLM for VrD-IE by separately incorporating the ISDR and GSDR information into the model. 

Tab. \ref{tab:gsdr} illustrates that both types of reading order information substantially benefits the downstream tasks. 
Since the method with ISDR behaves better in the majority of task settings, we choose to leverage the ISDR information in our main experiments.

\renewcommand\tabcolsep{5pt}
\begin{table}[t]
\small
\centering
\begin{spacing}{1.19}
\resizebox{1.01\linewidth}{!}{
\begin{tabular}{c|c|ccc}
\bottomrule
\textbf{Dataset} & \textbf{Task} & \textbf{None} & \textbf{ISDR} & \textbf{GSDR} \\
\hline
\multirow{2}{*}{\makecell[c]{FUNSD}} 
 & SER & 91.10 & \textbf{91.84} \textcolor{green2}{($\uparrow$0.74)} & 91.75 \textcolor{green2}{($\uparrow$0.65)} \\
 \cline{2-5}
 & EL & 88.06 & 88.46 \textcolor{green1}{($\uparrow$0.40)} & \textbf{88.59} \textcolor{green2}{($\uparrow$0.53)} \\
\hline
\multirow{2}{*}{\makecell[c]{EC-FUNSD}} 
 & SER & 83.62 & \textbf{84.34} \textcolor{green2}{($\uparrow$0.72)} & 84.13 \textcolor{green2}{($\uparrow$0.51)} \\
 \cline{2-5}
 & EL & 86.18 & \textbf{87.42} \textcolor{green3}{($\uparrow$1.24)} & 86.53 \textcolor{green1}{($\uparrow$0.35)} \\ 
\toprule
\end{tabular}
}
\end{spacing}
\vspace{-2mm}
\caption{
Ablation study on the impact of different types of reading order relation supervision. 
We report the performance of GeoLayoutLM on EC-FUNSD incorporating with different types of reading order information. }
\vspace{-4mm}
\label{tab:gsdr}
\end{table}

\section{Case Studies}
\label{sec:case}

\begin{figure*}[t]
    \centering
    \vspace{-4mm}
    \includegraphics[width=\linewidth]{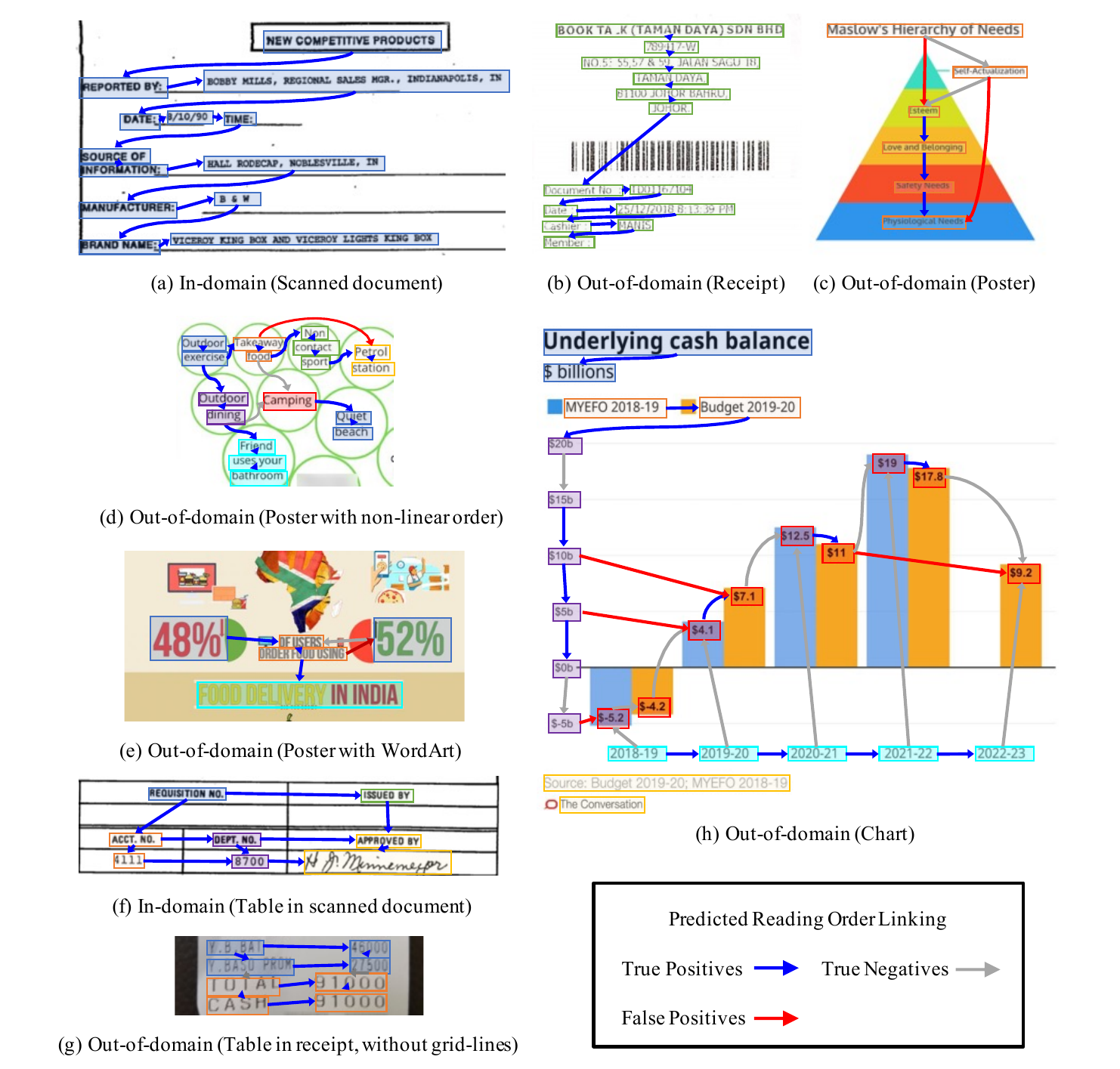}   
    \vspace{-4mm}
    \caption{
        Case study of the proposed reading order prediction model. 
        Each arrow represents a predicted relation linking between segments.
        }
    \vspace{-4mm}
    \label{fig:case-ro}
\end{figure*}

\begin{figure*}[t]
    \centering
    \vspace{-4mm}
    \includegraphics[width=\linewidth]{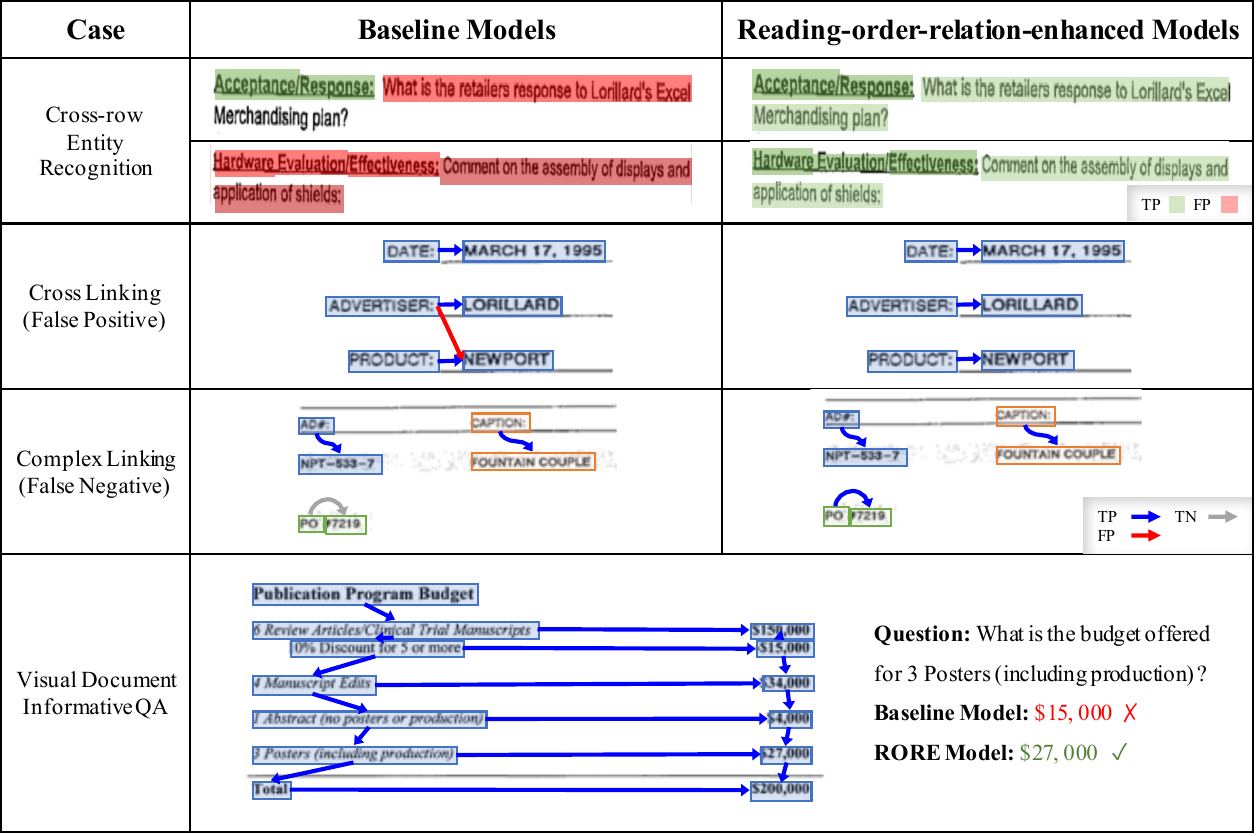}   
    \vspace{-4mm}
    \caption{
        Case study of baseline models and their corresponding reading-order-relation-enhanced variants for VrD-SER, VrD-EL and VrD-QA. 
        Entities are marked with shades and are distinguished by the shade of color. 
        Entity linking are marked with arrows.   
        }
    \vspace{-4mm}
    \label{fig:case-task}
\end{figure*}

\paragraph{Strengths and Weaknesses of the Proposed ROP Models}
For better understanding the strengths and weaknesses of the proposed ROP models, we conduct a case study for analyzing the prediction results by the LayoutLMv3-large model, both on in-domain datasets (e.g. FUNSD as scanned documents) and out-of-domain datasets (e.g. CORD as receipts, InfoVQA as posters). 
It is expected that the proposed model is able to recognize the reading pattern characteristic of in-domain samples, while also exhibiting some degree of generalization ability to out-of-domain samples.
Fig. \ref{fig:case-ro} visualizes the predict result of interested cases, revealing that: 
(1) \textbf{The proposed model has been acquainted with the major reading patterns of in-domain samples. }
According to Fig. \ref{fig:case-ro} (a-b), the proposed model links spatially-separated text regions which constitute a single content, while it also distinguishes irrelevant elements from the main content, such as page header and footer, seals, and watermarks. 
Besides, the proposed model performs well in identifying common non-linear reading order occasions in layouts, including table structures, as shown in Fig. \ref{fig:case-ro} (d, f-g).
It is also revealed that the proposed model does not exhibit a high false positive rate on table structures, unlike the other relation extraction models. 
We attribute the good phenomenon to the high annotation quality of the training data. 
(2) \textbf{The proposed model tends to over-rely on layout features, yet paying less attention to textual features. }
Fig. \ref{fig:case-ro} (c, e) depict that the model shows the tendency to predict a linking from an element to the nearest element on its right, disregarding their semantic correlation. 
In Fig. \ref{fig:case-ro} (h), confronted with a bar chart which is unseen during training, the model applies the pattern of processing tables, introducing false linking between horizontally-aligned elements.  
In Fig. \ref{fig:case-ro} (g), a single table is falsely recognized as two tables. The lower part of the table is recognized as a separate table since it has larger fonts and its bounding boxes are misaligned with the upper part. 
These sub-optimal outcomes suggest potential improvements of the proposed model to leverage the textual and vision features in a more efficient way. 

\paragraph{Positive Impact of RORE for Downstream VrD Tasks}
To determine the specific positive impact of RORE for downstream tasks, we conduct a case study by visualizing and comparing the prediction results of baseline models and their corresponding RORE variants for each task.
According to Fig. \ref{fig:case-task}, it reveals that the reading order relation inputs benefit the task by serving as a strong and reliable cue in the task. 
In EL and QA, the relationship between key/question entity and value/answer entity is highly overlapped with the successive relationship during reading, thus the latter information would be fully leveraged during training and serves as a key determinant in prediction. 
In SER, the reading order inputs hint the successive relationship between the words before and after the line break, which benefits the recognition of cross-row entities since RORE models would rely more on the reading order information than baseline models. 
These results also explain the success of utilizing pseudo reading order information in training RORE models, as moderate noise during training makes RORE models to focus on the more general reading order patterns that are also more robust in prediction, thus lead to steady performance improvement. 

\end{document}